\documentclass[journal]{IEEEtran}
\usepackage[T1]{fontenc}
\usepackage[utf8]{inputenc} 
\usepackage{newunicodechar}
\usepackage{orcidlink}

\usepackage{soul}
\newunicodechar{ }{ }
\newunicodechar{ε}{\ensuremath{\epsilon}}

\usepackage{graphicx}
\usepackage{amsmath}
\usepackage{amssymb} 
\usepackage{amsfonts}
\usepackage{multirow}
\usepackage{comment}
\usepackage[T1]{fontenc}
\usepackage[utf8]{inputenc}
\usepackage{xcolor}
\usepackage{colortbl}
\usepackage[utf8]{inputenc}
\usepackage[switch]{lineno}

\DeclareMathOperator{\atanTwo}{atan2}

\begin{document}
%
\title{Geo-Registration of Terrestrial LiDAR Point Clouds with Satellite Images without GNSS}

%
%
%


\author{%
    Xinyu~Wang$^{\dagger}$\,\orcidlink{0009-0008-4065-2472},~\IEEEmembership{Student Member,~IEEE,}
    Muhammad~Ibrahim\,\orcidlink{0000-0002-5376-2477},
    Haitian~Wang\,\orcidlink{0009-0000-7544-4667},~\IEEEmembership{Student Member,~IEEE,}
    Atif~Mansoor\,\orcidlink{0000-0001-6940-3914},
    Xiuping Jia\,\orcidlink{0000-0001-9916-6382},~\IEEEmembership{Fellow,~IEEE},
    and~Ajmal~Mian\,\orcidlink{0000-0002-5206-3842},~\IEEEmembership{Senior~Member,~IEEE}%
    \thanks{Xinyu Wang, Muhammad Ibrahim, Haitian Wang, Atif Mansoor, and Ajmal Mian are with the Department of Computer Science and Software Engineering, The University of Western Australia, Perth, WA 6009, Australia (e-mail: xinyu.wang@uwa.edu.au). Xiuping Jia is with the School of Engineering and Technology, University of New South Wales, Canberra, ACT 2612, Australia.\\
    Corresponding author: xinyu.wang@uwa.edu.au}

}

\maketitle
\begin{abstract}
 Accurate geo-registration of LiDAR point clouds remains a significant challenge in urban environments where Global Navigation Satellite System (GNSS) signals are denied or degraded. Existing methods typically rely on real-time GNSS and Inertial Measurement Unit (IMU) data, which require pre-calibration and assume stable signals. However, this assumption often fails in dense cities, resulting in localization errors. {\color{black}To address this, we propose a structured post-hoc geo-registration method that accurately aligns LiDAR point clouds with satellite images. The proposed approach targets point cloud datasets where reliable GNSS information is unavailable or degraded, enabling city-scale geo-registration as a post-processing solution.}
 Our method uses a pre-trained Point Transformer to segment road points, then extracts road skeletons and intersections from the point cloud and the satellite image. Global alignment is achieved through rigid transformation using corresponding intersection points, followed by local non-rigid refinement with radial basis function (RBF) interpolation. Elevation discrepancies are corrected using terrain data from the Shuttle Radar Topography Mission (SRTM). To evaluate geo-registration accuracy, we measure the absolute distances between the roads extracted from the two modalities. Our  method is validated on the KITTI benchmark and a newly collected dataset of Perth, Western Australia. On KITTI, our method achieves a mean planimetric alignment error of 0.69 m, corresponding to a 50\% reduction in global geo-registration bias compared to the raw KITTI annotations. On Perth dataset, it achieves a mean planimetric error of 2.17m from GNSS values extracted from Google Maps, corresponding to 57.4\% improvement over rigid alignment. Elevation correlation factor improved by 30.5\% (KITTI) and 55.8\% (Perth). A demonstration video is available at: \textcolor{blue}{\href{https://youtu.be/0wkACAB-O6E}{Demo Video}}.

\end{abstract}
\begin{IEEEkeywords}
LiDAR, Point Cloud, 3D City Map, Geo-registration, Skeletonization, Elevation Maps, Spatial Correction.
\end{IEEEkeywords}

%
\IEEEpeerreviewmaketitle

\vspace{-2mm}
\section{Introduction}


\IEEEPARstart{T}{he} rapid development of urban environments drives a growing demand for detailed 3D spatial data to support urban planning, infrastructure monitoring, and intelligent transportation \cite{URECH2020103903, Urabn12389, ijgi9090521, hu2020towards, wang2021navigation, zhou2021ndt}. Light Detection and Ranging (LiDAR) is the most prevalent technology for capturing high-resolution 3D point clouds, providing rich spatial representations of buildings, roads, vegetation, bridges, and other structures \cite{Urabn12389, MOKROS2021102512, 6942192}. 
This serves as a foundational data source for a wide range of geospatial analytics and decision-making tasks in complex urban environments~\cite{9855527, 7109840, GAO2023110862}.

Although significant advancements in point cloud modeling methods such as the Iterative Closest Point (ICP) algorithm variants \cite{zhang2024improved, koide2021voxelized, 10258436}, LiDAR Odometry and Mapping (LOAM) \cite{10.1109/IROS51168.2021.9636655, guo2022loam, oelsch2021r}, and Simultaneous Localization and Mapping (SLAM) \cite{10123040, 10973796} approaches have enabled the generation of city-scale models (e.g., Waymo  
\cite{sun2020scalability}, Perth CBD \cite{9647060} and Swan \cite{cg01-8n53-23} datasets), these are often generated without precise geographic referencing. This is largely due to limitations in the data acquisition process, where LiDAR systems collect data without a GNSS, or the GNSS and IMU sensor accuracy is compromised in urban environments \cite{guo2022loam, PAIJITPRAPAPORN2021100078}. Factors such as signal occlusion caused by tall buildings (the "urban canyon effect") \cite{groves2011shadow, wang2021safe}, sensor desynchronization \cite{antonopoulos2020sensor, degesys2007desync}, and inconsistent sampling rates between the sensors \cite{zhang2023radar} contribute to inaccurate or missing geographic metadata \cite{GAO2023110862}. Re-capturing data to fill in missing information is expensive and time consuming \cite{9196526, weinmann2013georeferenced} making post-hoc geo-registration a much more preferred alternative \cite{articleerror}. 

Precise georeferencing will allow the integration of point cloud data with standardized spatial coordinate systems, such as the World Geodetic System 1984 (WGS-84) and Universal Transverse Mercator (UTM) \cite{wang2024multi, groves2006accurate}. Without georeferencing, 3D city maps cannot be integrated with Geographic Information Systems (GIS), hindering multi-source spatial data fusion and comprehensive spatial analysis \cite{ijgi14050180, tang2022point, wang2024multi}. LiDAR point clouds often contain inherent spatial errors such as frame drift, global deformation, and loop closure failures \cite{ijgi14050180, articleerrors}. Without accurate geographic referencing, these errors cannot be quantified or effectively corrected, limiting the practical utility of 3D city models.

Several technical and environmental factors make precise geographic referencing of terrestrial LiDAR point clouds challenging. Firstly, most LiDARs such as Ouster, Velodyne, and RIEGL~VUX scanners, operate independently of GNSS, producing data in local coordinate frames \cite{article123895}. Secondly, significant incompatibilities exist between LiDAR and GNSS sensors, including differences in sampling frequency, data synchronization difficulties, and varying spatial resolution, resulting in misalignment \cite{GAO2023110862, PAIJITPRAPAPORN2021100078, article123895}. 
Furthermore, many existing point cloud datasets have missing GNSS information \cite{10123040, 9196526, hough-1}. 
Thus, there is a pressing need for a robust method that can accurately perform geo-registration of existing non-georeferenced point cloud data. 
{\color{black} To address these problems, this paper introduces a structured post-hoc geo-registration and spatial correction method based on topology-driven skeleton extraction, spatial transformations, and elevation correction, specifically targeting terrestrial LiDAR datasets that lack reliable geographic referencing.}
The proposed method begins with semantic segmentation to extract road points from raw LiDAR point clouds, which are then projected onto a plane to generate a simplified road skeleton. In parallel, high-resolution satellite maps are processed to produce matching road skeletons and intersection keypoints. Alignment is performed hierarchically: first, a rigid transformation aligns the two modalities using matched intersections, ensuring global consistency; then, a non-rigid refinement using radial basis function (RBF) interpolation corrects local distortions. Finally, terrain-aware elevation correction aligns the point cloud vertically using Shuttle Radar Topography Mission (SRTM) data, achieving full 3D spatial consistency. To comprehensively assess registration performance, we introduce two metrics that evaluate global consistency and local precision. 
%
%
Experiments are performed on the KITTI benchmark \cite{behley2019semantickitti} and a newly collected dataset in Perth city, Western Australia. 
On KITTI, our method reduces the mean planimetric alignment error from 1.38m to 0.69m (50\% improvement) and on Perth dataset, it achieves 2.17m against GNSS values extracted from Google Maps (57.4\% improvement). 
Elevation correlation improved by 30.5\%  on KITTI  and 50.4\% on the Perth dataset. 
A \textcolor{blue}{\href{https://youtu.be/0wkACAB-O6E}{Demo Video}} of the geo-registered 3D Perth city map shows our qualitative results.

The primary contributions of this research are as follows:

\begin{enumerate}
    {\color{black} \item We formulate and address a post-hoc city-scale geo-registration problem for terrestrial LiDAR point clouds, targeting practical scenarios where GNSS information is unavailable, unreliable, or erroneous during data acquisition.}

    \item We propose a cross-modal matching approach for LiDAR point clouds and satellite images by extracting shared geometric features in the form of skeleton representations comprising road centerlines and intersections.
    
    \item We propose a 
    geo-registration method that performs global rigid alignment via intersection matching, followed by local non-rigid refinement using radial basis functions and elevation correction with Shuttle Radar Topography Mission (SRTM) terrain data.

    

    \item We identify geo-registration errors in the KITTI benchmark \cite{behley2019semantickitti} and provide a corrected sub-dataset of the five longest KITTI sequences with refined geo-references.
    
    
\end{enumerate}


\vspace{-3mm}
\section{Literature Review}
\vspace{-1mm}
Accurate geo-registration of urban LiDAR point clouds remains a significant challenge, 
especially when GNSS is unavailable. Existing research can be divided into similar approaches, road feature extraction, 3D–2D registration, and non-rigid correction methods. These are detailed below.

Current approaches for GNSS coordinate annotation of point clouds primarily rely on multi-sensor fusion techniques that integrate GNSS with LiDAR and IMU data \cite{GAO2023110862,PAIJITPRAPAPORN2021100078} during collection. However, these methods face limitations in urban environments due to the "urban canyon effect” \cite{hough-1}, where signal occlusion degrades GNSS accuracy \cite{10123040}. Alternative approaches include 3D map-based global localization (MGL) using pre-existing 3D maps \cite{article123895}, which requires pre-constructed 3D point cloud maps for the corresponding mapping area. Additionally, SLAM-based methods that operate independently in GNSS-denied environments \cite{10.1109/IROS51168.2021.9636655,guo2022loam} have been developed. 
{\color{black}While these techniques achieve accurate relative positioning, they do not address the geo-registration of existing point cloud datasets collected without reliable GNSS to global geographic coordinates.}

Semantic segmentation and feature extraction from LiDAR point clouds have advanced through both classical and deep learning methods. Early works~\cite{rusu2011pcl, chen2009next} applied geometric and radiometric cues, while more recent studies employ neural architectures such as PointNet++~\cite{qi2017pointnetplusplus} and Point Transformer~\cite{wu2024ptv3} to segment road surfaces and urban structures. Skeletonization and thinning techniques~\cite{boyko2011extracting, yu2015learning} have been used to capture road centerlines and intersection topology, providing a concise and stable basis for spatial alignment.

Cross-modality registration approaches between 3D point clouds and 2D reference maps fall into projection-based and feature correspondence categories. Projection-based methods~\cite{gao2015automatic, Adaptive_bird} transform 3D point clouds into 2D representations for direct alignment with map imagery, but often discard useful geometric information. Feature-based techniques~\cite{schindler2012generation, wu2013voxel} extract salient features such as intersections or building footprints to establish correspondences across modalities. However, most require either coarse initial alignment or ground control points, which limits automation and scalability in large, GNSS-denied urban environments.

Rigid alignment corrects global misalignments but cannot address local geometric distortions inherent to mobile LiDAR mapping. RBF methods, such as thin-plate splines~\cite{bookstein1989principal, huang2013adaptive}, enable continuous non-rigid warping guided by control points, and have been applied in limited urban contexts~\cite{li2016reconstructing, ma2016non}. The selection of topological control points, such as road intersections, is critical to preserving spatial coherence during non-rigid correction.

Assessment of geo-registration accuracy without ground truth remains an open problem. Existing benchmarks focus primarily on small-scale or building-level datasets~\cite{dong2020registration, pomerleau2015review}. Quantitative metrics based on road centerline distances and intersection offsets~\cite{yu2014semiautomated, wu2019massive} offer practical alternatives for urban-scale evaluation but lack standardization across studies.

{\color{black} Existing approaches face key limitations for post-hoc geo-registration of terrestrial LiDAR point clouds. GNSS/IMU fusion and map-based global localization typically require reliable positioning information during data acquisition, while SLAM and LIO methods primarily focus on relative pose estimation and do not directly provide global geographic coordinates. Most existing 3D–2D registration approaches further assume coarse initial alignment or ground control points, limiting their applicability in large-scale urban environments. Moreover, rigid alignment alone cannot adequately correct the local distortions commonly observed in large-scale mobile mapping. These limitations highlight the need for a fully automatic framework that enables robust urban geo-registration by leveraging LiDAR data and publicly available satellite imagery, particularly in scenarios where reliable GNSS information is unavailable. Our work addresses these gaps with a unified approach that combines semantic road segmentation, skeleton-based topology matching, hierarchical rigid-to-non-rigid transformation, and terrain-aware elevation correction. }


\section{Methodology}
\vspace{-1mm}
\begin{figure*}
    \centering
    \includegraphics[height=0.44\textheight, width=1\linewidth]{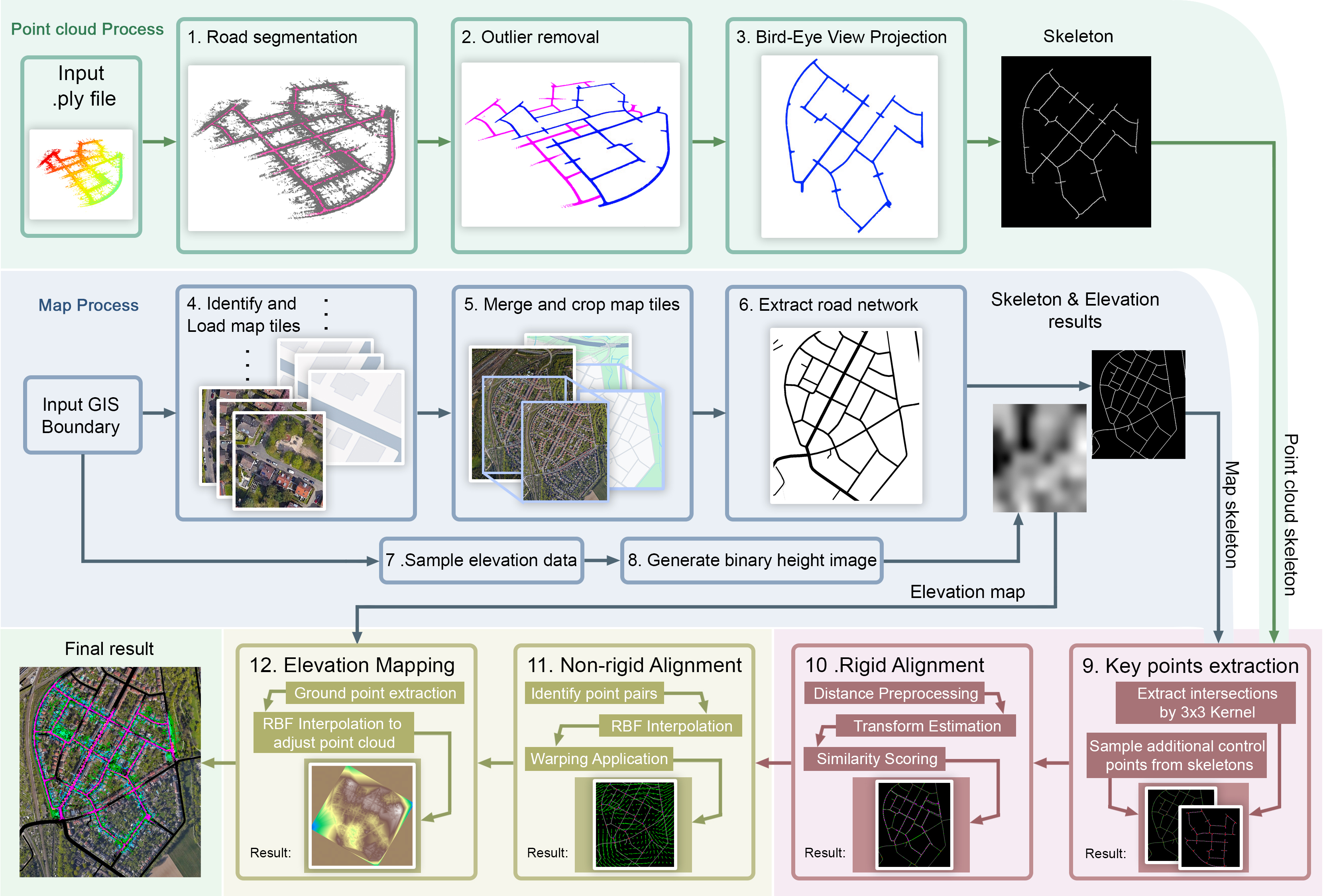}
     \vspace{-4mm}
    \caption{ \textbf{1}) Point cloud processing: road segmentation, outlier removal, and 2D projection to derive the road skeleton. \textbf{2}) Map processing: tile identification, loading, merging, cropping, road network extraction, and elevation sampling to generate skeleton and height maps. \textbf{3}) Rigid alignment: distance preprocessing, transform estimation, and similarity scoring to establish initial point cloud–map correspondence. \textbf{4}) Non-rigid alignment and elevation matching: point pair identification, RBF interpolation, warping, and ground point extraction to produce the final geo-registered point cloud. }
    \label{fig:pipeline}
    \vspace{-4mm}
\end{figure*}


Fig.~\ref{fig:pipeline} illustrates our method, which comprises four stages:
(1) preprocessing raw LiDAR point clouds to extract road surfaces and generate a skeleton representation; 
(2) extracting and refining reference map data to obtain a topologically consistent 2D road skeleton with elevation information; 
(3) hierarchical alignment, with rigid feature-based transformation for global correspondence and RBF-based non-rigid warping for local correction; and
(4) terrain-aware elevation adjustment to ensure vertical consistency with geographic reference data. 
Each stage is designed to address challenges in urban environments, including spatial noise, complex road topology, and discrepancies between map and point cloud domains. The following subsections detail each stage.

\vspace{-5mm}

\subsection{3D LiDAR Point Cloud Preprocessing}
Geo-registration of urban 3D point clouds without GNSS requires preprocessing to isolate road surfaces and derive a skeleton suitable for map alignment. The pipeline consists of five stages: semantic segmentation, adaptive voxel downsampling, local outlier removal, disconnected cluster filtering, and skeletonization via 2D projection.

\noindent\textbf{Semantic segmentation.} 
Road surface points are extracted using Point Transformer V2~\cite{wu2022point}, which employs vector self-attention to capture local geometric relationships. For a point $x_i$, the attention is defined as:
\begin{equation}
\mathrm{VSA}(x_i) = \sum_{j \in N(i)} \rho(\phi(x_i) - \psi(x_j) + \delta_{ij}) \odot \gamma(x_j),
\end{equation}
where $N(i)$ denotes neighbors of point $i$; $\phi$, $\psi$, and $\gamma$ are learnable mappings; $\delta_{ij}$ encodes relative position; $\rho$ is a nonlinear normalization; and $\odot$ is element-wise multiplication. This enables accurate road–non-road separation by modeling spatial context.

\noindent\textbf{Adaptive voxel downsampling.} 
To reduce computational cost while preserving geometry, the segmented road cloud is adaptively voxelized. The voxel size is defined as $s = \alpha \cdot \left( \frac{\mathrm{vol}(P)}{|P|} \right)^{1/3}$, where $\mathrm{vol}(P)$ is the bounding-box volume, $|P|$ the point count, and $\alpha$ a scaling factor. {\color{black} In all experiments, $\alpha$ is set to 1.1, such that the voxel size is slightly larger than the average point spacing.} One representative point per voxel is retained, substantially reducing density while maintaining structural fidelity.

\noindent\textbf{Local outlier removal.} 
Residual noise and isolated points can degrade downstream processing. To suppress them, we apply statistical outlier removal: for each point $p_i$, the mean distance $\mu_i$ and standard deviation (std) $\sigma_i$ to its $k$ nearest neighbors ($k=20$) are computed. A point is an inlier if $(\mu_i + \beta \sigma_i) < \tau(p_i)$, where $\beta$ is a fixed threshold (2.5) and $\tau(p_i)$ is an adaptive distance threshold based on local density. This removes points that deviate from the spatial distribution of their neighborhood.

\noindent\textbf{Density-based cluster filtering.} 
After outlier removal, small disconnected clusters often remain due to occlusions or noise. We apply Density-Based Spatial Clustering of Applications with Noise (DBSCAN)~\cite{10.5555/3001460.3001507} and retain only dominant road clusters. For each point $p_i$, the neighborhood radius is adaptively set as $\epsilon_i = \gamma \cdot \tfrac{1}{k} \sum_{j=1}^{k} \|p_i - p_j\|$, where $\gamma=1.8$ and $k$ is the number of neighbors. Clusters with fewer than $\mathrm{minPts}=17$ points are discarded. This eliminates fragmented or low-density regions, further refining the road structure.


\noindent\textbf{2D projection and skeletonization.} 
The denoised road cloud is projected onto the $XY$-plane to form a 2D representation. Morphological thinning is then applied to extract the centerline skeleton $S$, producing a single-pixel-wide structure that preserves connectivity, including intersections. Spurious short branches are pruned as
\vspace{-1mm}
\begin{equation}
S_\text{pruned} = S \setminus \{p \in S \mid d_\text{endpoint}(p) < L_\text{min} \land \kappa(p) < \kappa_\text{th}\},
\end{equation}
where $d_\text{endpoint}(p)$ is the distance from point $p$ to the closest skeleton end point, $\kappa(p)$ is local curvature, $L_\text{min}$ is minimum branch length, and $\kappa_\text{th}$ is curvature threshold. This results in a robust topological graph (i.e. a skeleton) for subsequent map alignment.

\vspace{-3mm}

\subsection{2D Reference Map Preprocessing}
Accurate geo-registration requires that both the point cloud and the reference map share a comparable, topologically consistent road skeleton. To this end, we process high-resolution maps to extract a 2D skeletonized road structure similar to the one extracted from the point cloud. 

\noindent\textbf{Acquisition of reference maps.} Georeferenced satellite imagery and vector road data are obtained for each region using the Google Maps API in Mercator projection, ensuring uniform resolution and a consistent coordinate system. Raw map data is not directly usable for alignment due to stylized road widths, simplified semantics, and occasional topological inconsistencies.

\noindent\textbf{Color-based segmentation.} 
To derive an initial road mask, the RGB map is converted to HSV space and thresholded to isolate bright, low-saturation road regions. For a pixel $(x,y)$, the mask is defined as $M_\text{init}(x,y) = 1$ if $H_{\min} \leq H(x,y) \leq H_{\max}$, $S(x,y) \leq S_{\text{th}}$, and $V(x,y) \geq V_{\text{th}}$, and $0$ otherwise, where $H,S,V$ denotes the hue, saturation and value channels. Thresholds $(H_{\min}, H_{\max}, S_{\text{th}}, V_{\text{th}})$ are empirically set per region based on road color distributions.

\noindent\textbf{Bidirectional connected component filtering.} 
The initial mask includes isolated noise, fragmented segments, and small holes due to imaging artifacts. We apply connected component analysis to the roads ($S_i$) and background ($B_j$) regions, and remove components with an area below a fraction $\beta$ of the mean size of the road component $\bar{S}$. The filtered mask is given by $M_{\text{filtered}} = \bigcup_{i: |S_i| \geq \beta \bar{S}} S_i \;\cup\; \bigcup_{j: |B_j| \geq \beta \bar{S}} B_j$, where $\beta$ controls the minimum preserved size and $|\cdot|$ denotes component area in pixels. This dual filtering removes small artifacts and fills minor gaps, preserving the dominant road network topology.

\noindent\textbf{Skeletonization and topology refinement.} 
To obtain a road skeleton suitable for topological matching, we apply morphological thinning to the filtered road mask. This produces a single-pixel-wide, connectivity-preserving structure. Spurious branches are then pruned by evaluating their length $L_b$ and mean curvature $\kappa_b$; branches with $L_b < L_{\min}$ and $\kappa_b < \kappa_{\text{th}}$ are removed. The refined skeleton is
$S^*_\text{map} = S_\text{map} \setminus \bigcup_{b \in \mathcal{B}} \{ b : L_b < L_{\min} \wedge \kappa_b < \kappa_{\text{th}} \}$,
where $\mathcal{B}$ is the set of skeleton branches, $L_{\min}$ the minimum length, and $\kappa_{\text{th}}$ the curvature threshold.

\vspace{-3mm}

\subsection{Rigid alignment based on feature extraction}
Aligning point clouds with reference maps is challenging due to differences in dimensionality, scale, and abstraction. We address this by extracting repeatable features from both modalities and estimating a rigid transformation to ensure global and local spatial consistency.

\noindent\textbf{Keypoint detection.} 
From skeletonized road networks, three types of keypoints are extracted:  
(1) \textit{Intersection points}, where multiple roads meet. For each pixel $p \in S_{pc}$, if its $3\times3$ neighborhood $\mathcal{N}_{3\times3}(p)$ contains more than two skeleton branches, then $p$ is classified as an intersection point, i.e., $I_{pc} = \{p \in S_{pc} : |\mathcal{N}_{3\times3}(p) \cap S_{pc}| > 2\}$.  
(2) \textit{Control points}, sampled at regular intervals (density = 20) along skeleton branches to capture elongated segments.  
(3) \textit{Corner points}, corresponding to the geometric boundary of the skeletonized image, providing global orientation cues. The union of these keypoints captures both topology and geometry, enabling stable correspondence matching.

\noindent\textbf{Rigid transformation estimation.} 
Given keypoint sets $\mathcal{P}_1$ (point cloud) and $\mathcal{P}_2$ (map), pairwise matches are used to estimate a similarity transformation (scale, rotation, translation). For point pairs $(p_i^1, p_k^1) \in \mathcal{P}_1$ and $(p_j^2, p_l^2) \in \mathcal{P}_2$, the scale is $s = \|p_k^1 - p_i^1\| / \|p_l^2 - p_j^2\|$. The rotation angle is:

{\color{black}

\begin{equation}
\begin{split}
\theta = -\Big[
&\atanTwo(p_k^1[1] - p_i^1[1],\, p_k^1[0] - p_i^1[0]) \\
&- \atanTwo(p_l^2[1] - p_j^2[1],\, p_l^2[0] - p_j^2[0])
\Big]
\end{split}
\end{equation}

}

and the similarity transformation applied to $p \in \mathcal{P}_2$ is:
\begin{equation}
    \mathcal{T}'(p) = s \cdot \mathbf{R}_\theta (p - p_j^2) + p_i^1 + \mathbf{t}
\end{equation}
where $\mathbf{R}_\theta$ is the 2D rotation matrix and $\mathbf{t}$ is a translation vector.

\noindent\textbf{Selecting the optimal global alignment.} 
Each candidate transformation is evaluated using a proximity-based matching score. For every transformed keypoint, the minimal Euclidean distance to the corresponding set is computed, and the score is:
\begin{equation}
\begin{split}
    S = \alpha \sum_{i=1}^{|\mathcal{P}_1|} \mathbb{I} \Big(\min_{j} \|p_i^1 - \mathcal{T}'(p_j^2)\| < \epsilon \Big) \\
    + \beta \sum_{i=1}^{|\mathcal{A}_1|} \mathbb{I} \Big(\min_{j} \|a_i^1 - \mathcal{T}'(a_j^2)\| < \epsilon \Big)
\end{split}
\label{eq:score}
\end{equation}
where $\mathbb{I}$ is the indicator function, $\epsilon$ the distance threshold, and $\alpha,\beta$ the weights for primary and auxiliary keypoint sets. The optimal parameters \( (s^*, \theta^*, \mathbf{t}^*) \) maximize the alignment score, given by \( (s^*, \theta^*, \mathbf{t}^*) = \arg\max_{s, \theta, \mathbf{t}} S(s, \theta, \mathbf{t}) \).

\noindent\textbf{Global point cloud transformation.} 
Using the optimal parameters, the entire point cloud is aligned to the reference map through centering, rescaling, and rigid transformation. Each point $\mathbf{p}$ is centered as $\mathbf{p}_c = \mathbf{p} - \mathbf{c}$, where $\mathbf{c}$ is the centroid, and the aligned point is computed as $\mathbf{p}' = \mathbf{R}_{\theta^*}(s^* \mathbf{p}_c) + \mathbf{c} - \mathbf{t}^*$, with $s^*$, $\mathbf{R}_{\theta^*}$, and $\mathbf{t}^*$ denoting the optimal scale, rotation, and translation. This transformation ensures geometric consistency and provides a robust initialization for non-rigid refinement.

\vspace{-2mm}




\subsection{Non-rigid transformation using radial basis functions}
Rigid alignment provides initial correspondence between the point cloud and the map, but does not correct local distortions in large-scale urban data. To address these residual misalignments, we apply a non-rigid transformation based on radial basis function (RBF) interpolation, enabling smooth warping of the point cloud to the reference map.

\noindent\textbf{Construction of the deformation field.} 
Let $\{\mathbf{s}_i\}_{i=1}^n$ and $\{\mathbf{t}_i\}_{i=1}^n$ be matched keypoints from the point cloud and map skeletons. The deformation field is defined as
\begin{equation}
    \mathbf{f}(\mathbf{x}) = \sum_{i=1}^{n} \mathbf{w}_i \phi(\|\mathbf{x} - \mathbf{s}_i\|) + \mathbf{P}(\mathbf{x}),
\end{equation}
where $\mathbf{w}_i$ are weights determined by the displacements $\mathbf{t}_i - \mathbf{s}_i$, 
$\phi(\cdot)$ is the radial basis kernel, and $\mathbf{P}(\mathbf{x})$ is a polynomial term for global affine adjustment. 
We adopt the thin-plate spline (TPS) kernel, defined as {\color{black}$\phi(r)=r^2\ln(r)$.}
{\color{black}In implementation we set $\phi(0)=0$ by continuity, and the previously shown multiquadric form $\sqrt{r^2+\epsilon^2}$ is corrected here for consistency with TPS.}


\noindent\textbf{Control point selection and correspondence.} 
The corresponding control points are established through a search for the nearest neighbor within a spatial threshold $\tau = 30$:
\begin{equation}
    \mathcal{C} = \left\{ (\mathbf{s}_i, \mathbf{t}_i) \mid \mathbf{t}_i = \arg\min_{\mathbf{t} \in \mathcal{T}} \|\mathbf{s}_i - \mathbf{t}\|,\, \|\mathbf{s}_i - \mathbf{t}_i\| < \tau \right\}.
\end{equation}
The resulting displacement field is computed separately for $x$ and $y$ directions by fitting the RBF interpolators to the displacements of the control points.

\noindent\textbf{Application of non-rigid transformation.}
The deformation field $\mathbf{f}(\mathbf{x})$ is applied to all point cloud coordinates, correcting local geometric distortions while preserving global structure. This yields a spatially coherent alignment consistent with both large-scale and fine-grained features of the reference map.






\vspace{-3mm}
\subsection{Terrain-aware elevation matching}
Accurate vertical alignment is required to integrate the registered point cloud into geographic information systems and to preserve geometric fidelity across varied terrain. After horizontal registration, we apply a terrain-aware correction to reconcile point elevations with reference terrain heights.

\noindent\textbf{Ground point extraction and grid-based modeling.} 
Ground points are segmented using RANSAC plane fitting with an initial threshold $\tau_h = 0.05$, iteratively refined to $\tau_{\text{min}} = 0.02$. For each grid cell $(x,y)$ at resolution $r=1.0$, the mean ground elevation is computed as $h(x,y) = \tfrac{1}{|S_{x,y}|}\sum_{p \in S_{x,y}} z_p$, where $S_{x,y}$ is the set of ground points in the cell and $z_p$ their elevations.

\noindent\textbf{Integration with reference terrain data.}  
Absolute terrain heights from the reference map are used to assign global elevation values. Each ground point is adjusted to the terrain height at $(x,y)$, while non-ground points (e.g., buildings, vegetation) retain their relative height by adding the original offset above the local ground surface.



\noindent\textbf{Outlier suppression and local consistency.}  
To ensure elevation consistency, a statistical filter is applied to the grid-based model. Points are classified as outliers when
\begin{equation}
\mathcal{O} = \{\, p \in G : |z_p - \mu_{S_{x,y}}| > 2.5 \cdot \max(\sigma_{S_{x,y}}, \tau_{\text{min}}/2) \,\},
\end{equation}
where $\mu_{S_{x,y}}$ and $\sigma_{S_{x,y}}$ are the mean and standard deviation of cell heights. This filtering removes artifacts while preserving a terrain profile consistent with the reference surface.

\vspace{-2mm}
\subsection{Registration evaluation metrics}
To assess geo-registration accuracy, we use two metrics based on distance errors between corresponding road centerline points and intersections across the two modalities. Since the map is scaled to the point cloud (in meters), errors are reported in absolute units. A key advantage is that these metrics do not require GNSS ground truth and also provide elevation error estimates using NASA’s SRTM30 data.

\noindent\textbf{Road centerline distance analysis.}
For each trajectory point $p_i \in \mathbb{R}^2$ in the registered point cloud, we compute its shortest distance to the nearest road centerline segment $s_j = (\mathbf{a}_j, \mathbf{b}_j)$, with endpoints $\mathbf{a}_j, \mathbf{b}_j \in \mathbb{R}^2$. The segment direction is $\vec{v}_j = \mathbf{b}_j - \mathbf{a}_j$, and the perpendicular distance to segment $s_{j^*}$ is
\begin{equation}
d_{\perp}(p_i, s_{j^*}) =
\frac{\left\| \vec{v}_{j^*} \times \left(p_i - \mathbf{a}_{j^*}\right) \right\|}
     {\|\vec{v}_{j^*}\|},
\end{equation}
where $s_{j^*}$ is the segment minimizing $d_{\perp}(p_i, s_j)$ under orthogonal projection. If the projection lies outside the segment, the endpoint distance is used:
\begin{equation}
d(p_i, s_j) =
\begin{cases}
d_{\perp}(p_i, s_j), & \text{if } \lambda^* \in [0,1], \\
\min(\|p_i - \mathbf{a}_j\|, \|p_i - \mathbf{b}_j\|), & \text{otherwise},
\end{cases}
\end{equation}
with $\lambda^* = \frac{(p_i - \mathbf{a}_j)\cdot \vec{v}_j}{\|\vec{v}_j\|^2}$ the normalized projection position.  

This produces distances $\mathcal{D}=\{d_i\}_{i=1}^N$ for $N$ trajectory points. Since all distances are nonnegative, their distribution follows a half-normal form. To enable parametric inference, we symmetrize the data as $\mathcal{D}_{\text{sym}} = \{d_i\}_{i=1}^N \cup \{-d_i\}_{i=1}^N$ and fit a normal distribution via maximum likelihood estimation (MLE):
\begin{equation}
(\hat{\mu}, \hat{\sigma}) = \arg\max_{\mu,\sigma} \prod_{i=1}^{2N} \frac{1}{\sigma\sqrt{2\pi}}
\exp\!\left(-\frac{(d_i-\mu)^2}{2\sigma^2}\right).
\end{equation}

The normal model provides a concise measure of alignment dispersion. The estimated mean $\hat{\mu}$ and standard deviation $\hat{\sigma}$ characterize average offset and spread, with smaller $\hat{\sigma}$ indicating tighter concentration around the centerline. Local misalignments are flagged using the threshold $\tau = \hat{\mu} + 2\hat{\sigma}$, classifying any $d_i > \tau$ as a significant deviation.



\noindent\textbf{Intersection offset evaluation.}
Intersection centroids are extracted from both the registered point cloud and the reference map. Let $\mathcal{P}_I = \{p_i\}$ and $\mathcal{M}_I = \{m_i\}$ denote the sets of intersection centers, with one-to-one correspondences established by nearest-neighbor search within radius $\delta$. The mean intersection offset is defined as $E_{\text{int}} = \tfrac{1}{K} \sum_{i=1}^{K} \|p_i - m_i\|_2$, where $K = |\mathcal{P}_I|$ is the number of matched pairs. This metric captures the spatial alignment of salient topological features. Since intersections are structurally stable and visually distinct across modalities, the measure remains robust even without GNSS or dense ground truth, reducing the sensitivity to local geometric noise.

\section{Experiments}
We describe the acquisition of the KITTI and Perth LiDAR datasets, followed by the experimental setup, including hardware, software, and algorithmic parameters. Results are reported for both planimetric (XY) and elevation (Z) alignment. Evaluation metrics include centerline distance error, GNSS reference error (KITTI only), and elevation differences and correlation with SRTM30 data.

\vspace{-2mm}
\subsection{Dataset Description}
Here we describe the KITTI and Perth datasets as well as the geo-referenced (Google) map and SRTM30 elevation data.



\noindent\textbf{KITTI dataset.}
The KITTI benchmark~\cite{Geiger2013IJRR} was collected in Karlsruhe, Germany, and includes synchronized data from a Velodyne HDL-64E LiDAR, high-resolution stereo cameras, and a GPS/IMU navigation system. The LiDAR produces 3D point clouds at 10~Hz, with up to 130,000 points per scan and a 360° field of view. The GPS/IMU subsystem provides centimeter-level pose annotations for each frame. For our experiments, we selected sequences 00, 02, 05, 08, and 09, which together span diverse urban settings including downtown areas, residential neighborhoods, and suburban roadways. These sequences exhibit complex road geometries and challenges such as occlusions, dynamic objects, and varying surfaces. Each sequence includes several thousand consecutive frames with associated GNSS information. Our method later reveals that these annotations are inaccurate in urban environments.

\noindent\textbf{Perth CBD LiDAR Map 2021.}
This dataset \cite{s2p2-2e66-23} is a high-resolution 3D map of Perth city in Western Australia. LiDAR data collection and map construction were both performed by our team. Data acquisition was performed using a vehicle-mounted Ouster LiDAR with 64 vertical channels and up to 2048 planimetric scan resolution. The sensor was mounted 1.5~m above ground level and collected 10 frames per second. Data collection spanned three two-hour sessions, covering all major roads in a closed-loop fashion to minimize drift and maximize map coverage. The resulting dataset comprises approximately 150,000 frames, totaling over 64 million points and encompassing a ground area of approximately $3~\mathrm{km} \times 1.2~\mathrm{km}$ (nearly $4~\mathrm{km}^2$). This dataset does not contain GNSS or IMU information, making it an ideal testbed for geo-registration methods operating under GNSS-denied conditions. 

\noindent\textbf{Reference Map and Elevation Data.}
For both datasets, 2D reference maps were sourced from the Google Maps API at zoom level 18, with Mercator projection. This projection method is consistent with Google Maps GNSS coordinate system. At the Perth dataset's latitude ($-31^\circ$), the map resolution is $0.512$ meters per pixel and at the KITTI dataset's latitude ($49^\circ$), the resolution is $0.392$ meters per pixel. For vertical alignment, we use the SRTM 30m Global 1 arc second V003 dataset (NASA/NGA) \cite{srtm_v003}, available for both cities (Perth and Karlsruhe) and provides elevation values at a spatial resolution of 30m and vertical RMSE of $\pm3.56$ meters. 

\begin{table}[t]
\centering
\caption{Planimetric (xy plane) alignment errors on the KITTI dataset.}
\label{tab:kitti_xy_acc}
\vspace{-3mm}
\footnotesize{\textit{Note: Meter values are reported as mean $\pm$ std. Our alignment (shaded column) is more accurate than the GNSS values in the KITTI dataset.} \vspace{1mm}}
\begin{tabular}{c|c>{\columncolor{yellow!40}}cc}
\hline
Seq. & Before Correction & After Correction & KITTI GNSS \\
\hline
00 & $2.11 \pm 2.60$ & $0.63 \pm 0.76$ & $0.89 \pm 1.22$ \\
02 & $2.49 \pm 3.42$ & $0.77 \pm 0.88$ & $1.03 \pm 1.45$ \\
05 & $1.66 \pm 2.10$ & $0.45 \pm 0.57$ & $1.13 \pm 1.45$ \\
08 & $4.26 \pm 5.21$ & $1.08 \pm 1.35$ & $2.14 \pm 2.95$ \\
09 & $2.48 \pm 2.90$ & $0.53 \pm 0.64$ & $1.72 \pm 2.35$ \\
\hline
Average & $2.60 \pm 3.25$ & $0.69 \pm 0.84$ & $1.38 \pm 1.88$ \\
\hline
\end{tabular}%
\vspace{-3mm}
\end{table}

\vspace{-2mm}
\subsection{Experimental Setup}

All experiments were performed on a workstation with Intel Core i9-10900K (3.7 GHz), 32~GB DDR4 memory, and an NVIDIA RTX 3080 GPU (12~GB VRAM). 
Implementation was done in Python 3.9 using Anaconda environment. Core libraries included NumPy, SciPy, Open3D, scikit-learn, Matplotlib, and OpenCV 4.5. Deep learning inference for semantic segmentation was based on PyTorch 1.13 with CUDA 11.6 support. Algorithm parameters were chosen as follows: keypoint matching threshold = 8 pixels, skeleton extraction density =  40, RBF interpolation parameter $\epsilon=5$, and correspondence distance threshold = 30 pixels. Point cloud preprocessing used 20 nearest neighbors and a std multiplier of 2.0 for outlier removal. 
\vspace{-3mm}


\vspace{-2mm}
\subsection{Results on the KITTI Dataset}

We report both planimetric (XY plane) and elevation (Z axis) alignment errors (mean absolute value in meters $\pm$std) for the five longest sequences of KITTI dataset. 

\begin{figure*}[htbp]
    \centering
    \includegraphics[width=1\linewidth]{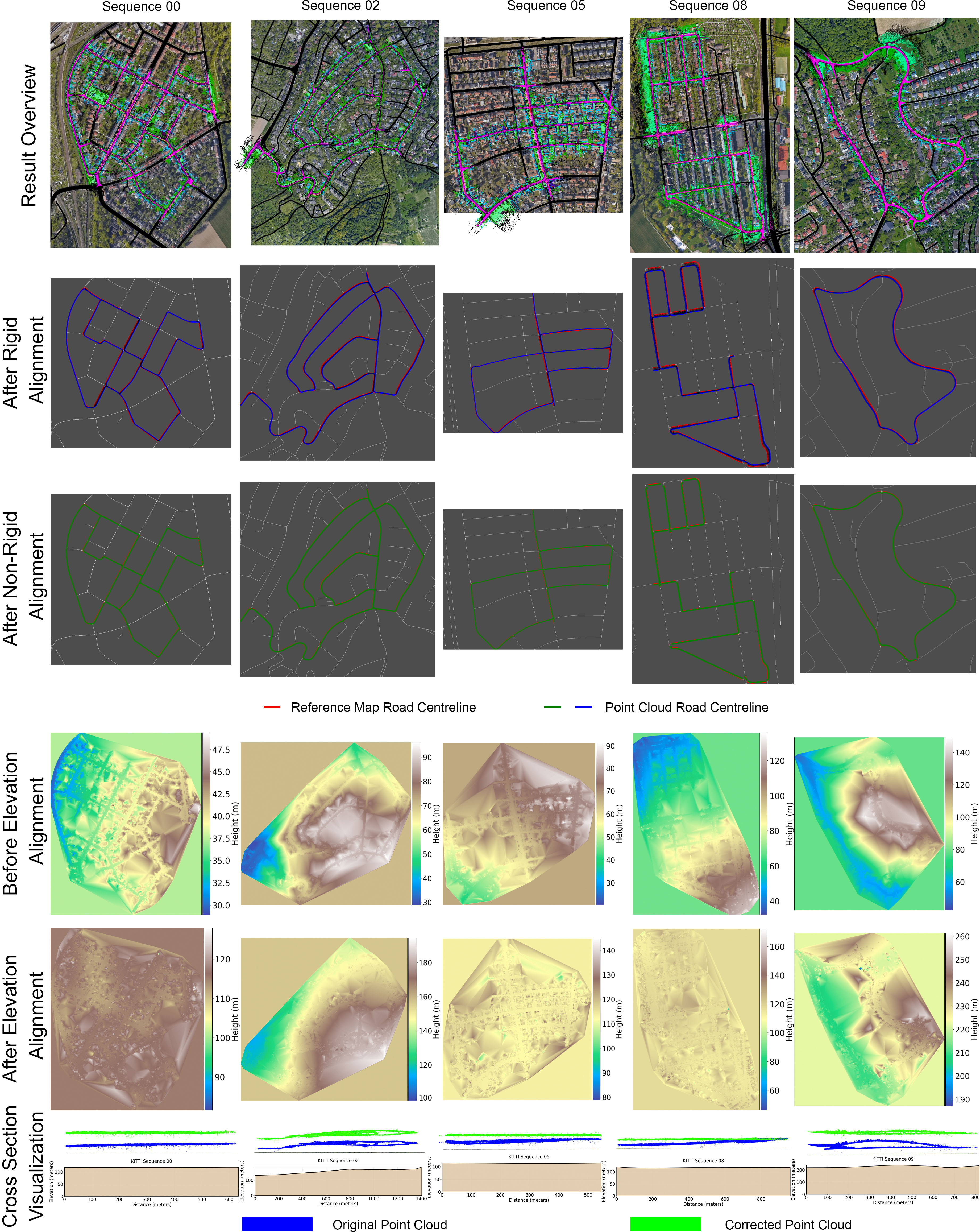}
    \caption{Alignment results for selected KITTI sequences (00, 02, 05, 08, 09). Magenta curves show geo-registered point cloud trajectories overlaid on reference map tiles, illustrating correspondence with the urban road network.}

    \label{fig:kitti_overall}
    \vspace{-4mm}
\end{figure*}

\noindent\textbf{Planimetric Alignment on KITTI Dataset.}
%
Fig.~\ref{fig:kitti_overall} (top) shows alignment between the LiDAR trajectory and the road map. The registered point cloud trajectories (magenta) are superimposed on reference map backgrounds. We can see that the registered trajectories accurately follow the underlying road map, including at intersections and curved regions, without GNSS-based initialization.


\begin{table}[t!]
\centering
\caption{Planimetric alignment errors on the Perth dataset.}
\label{tab:perth_xy_acc}
\vspace{-3mm}
\footnotesize{\textit{Note: Meter values are reported as mean $\pm$ std. Our alignment (shaded column) shows significant improvement over the before correction values.}\vspace{1mm}}
\begin{tabular}{c|c>{\columncolor{yellow!40}}c}
\hline
Seq. & Before Correction & After Correction \\
\hline
01 & $7.32 \pm 4.34$ & $0.87 \pm 0.95$ \\
02 & $3.03 \pm 3.19$ & $0.83 \pm 0.93$ \\
03 & $1.97 \pm 2.18$ & $0.84 \pm 0.89$ \\
Merged & $5.09 \pm 7.27$ & $2.17 \pm 1.07$ \\
\hline
\end{tabular}%
\vspace{-5mm}
\end{table}

\begin{figure*}
    \centering
    \includegraphics[height=0.5\textheight, width=1\linewidth]{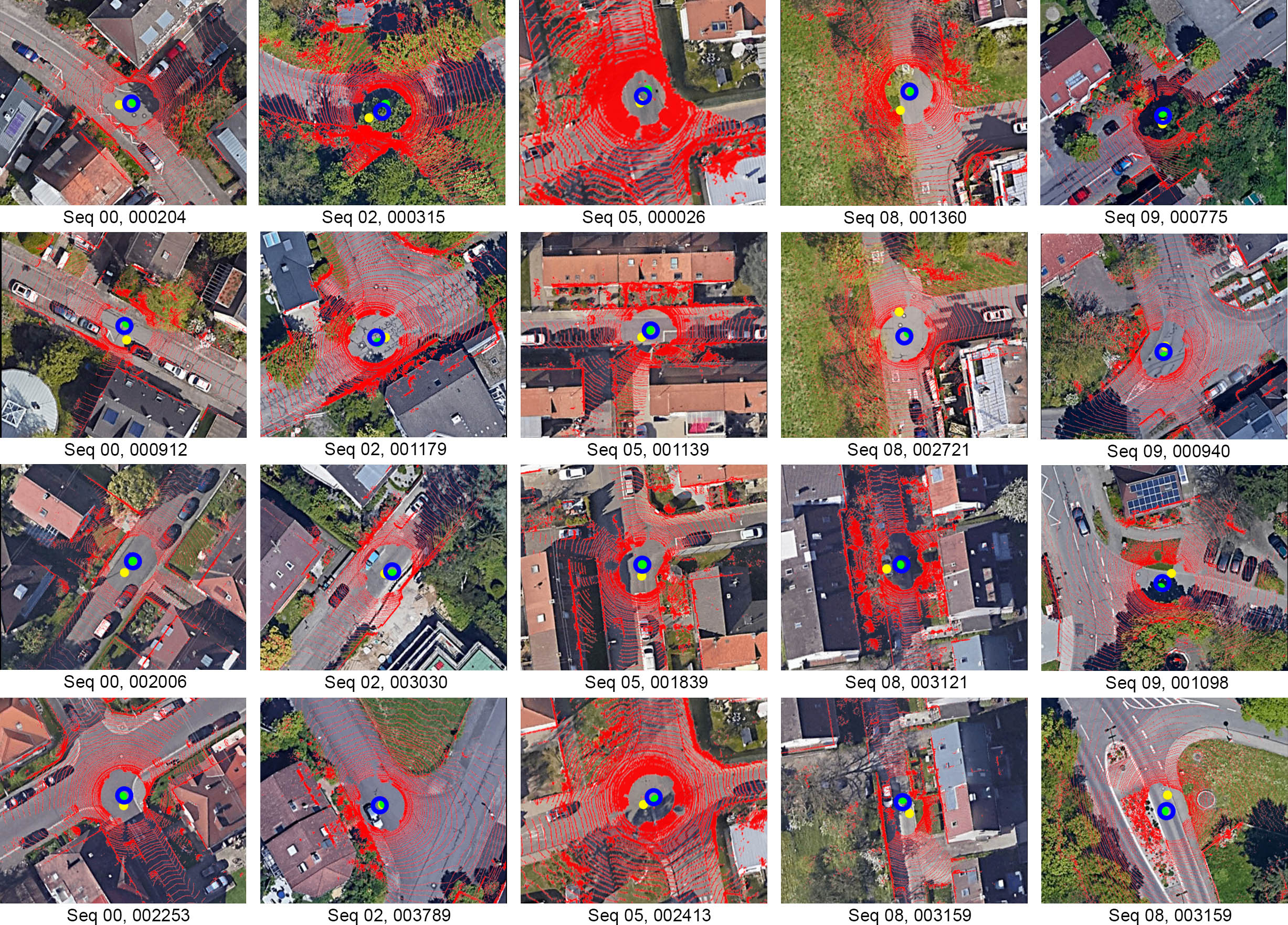}
    \caption{Frame-level registration results for selected frames in each KITTI sequence. Green dots: estimated trajectory points after geo-registration. Blue dots: manually labeled ground truth. Yellow dots: GPS-based reference. LiDAR points (red) are projected on the underlying map imagery. The results confirm accurate local alignment under diverse urban conditions.}
    \vspace{-4mm}
    \label{fig:Grid_vis_kitti}
\end{figure*}

Fig.~\ref{fig:Grid_vis_kitti} shows frame-level alignment results for some samples from each of the five KITTI sequences. In each frame, the {green dot} is the estimated position after geo-registration using our method, the {blue circle} indicates the actual center of the point cloud, and the {yellow dot} shows the original GNSS location provided in the raw KITTI data. The {red overlay} corresponds to the LiDAR point cloud projected onto the map. We can see strong consistency between the {green} dots and {blue} circles, demonstrating that our method is more accurate than the GNSS locations provided in the raw KITTI data. 

We also compute the mean absolute Euclidean distances between the registered LiDAR trajectory and the corresponding road centerline in the reference map. 

\begin{figure*}
    \centering
    \includegraphics[width=1\linewidth]{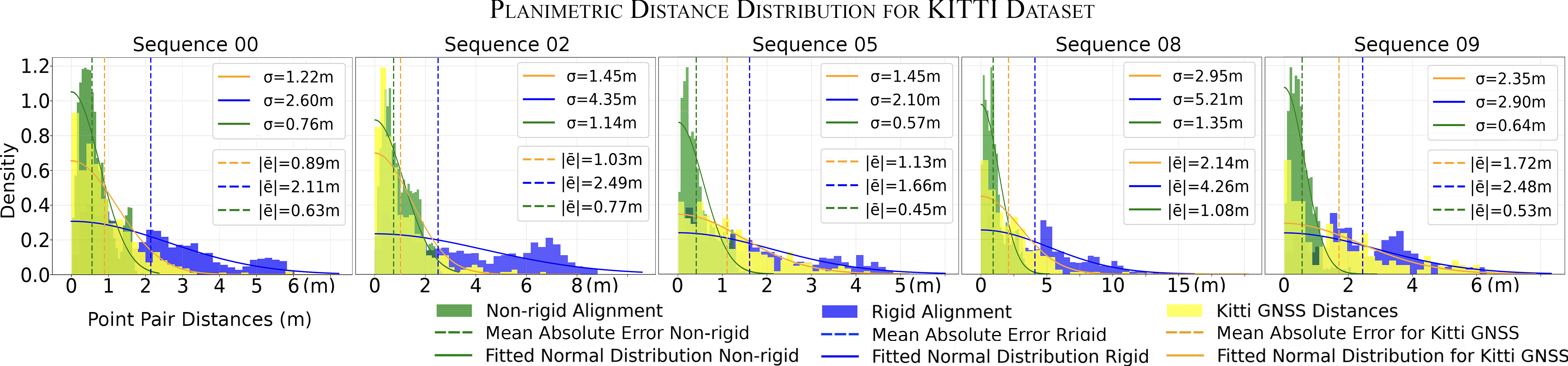}
    \vspace{-5mm}
    \caption{Point-pair distance distribution analysis across five KITTI sequences. Histograms compare distances between KITTI GNSS data (yellow) and our point cloud method using rigid alignment (blue) and non-rigid alignment (green), with fitted normal distributions and std ($\sigma$) for each approach.}
    \label{fig:kitti_dist00}
    \vspace{-2mm}
\end{figure*}

\begin{figure*}
    \vspace{-3mm}
    \centering
    \includegraphics[width=1\linewidth]{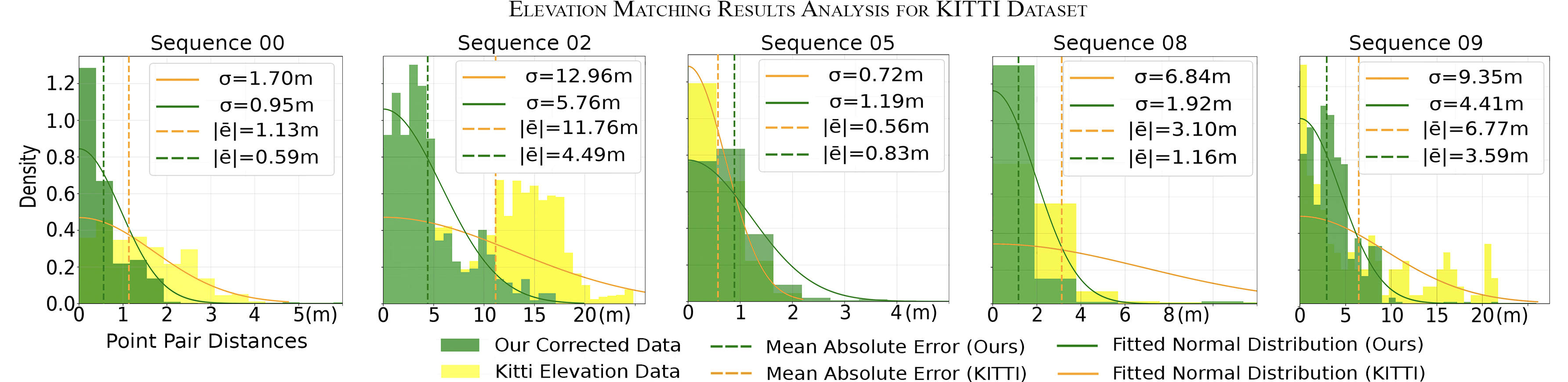}
    \vspace{-5mm}
    \caption{Cross-sequence validation of elevation correction accuracy. Comparison of the proposed point cloud-based method with KITTI GNSS ground truth across sequences 00, 02, 05, 08, and 09. The left plot shows index-based correlations between corrected and ground truth elevations, and the right plot presents error probability distributions with fitted std ($\sigma$). The bottom-left subfigure visualizes corrected elevations in the XY-plane for sequence~09.}

    \label{fig:kitti_z00}
    \vspace{-5mm}
\end{figure*}

Table~\ref{tab:kitti_xy_acc} summarizes our results, reporting alignment errors before and after correction with non-rigid transformation. For comparison, raw KITTI GNSS errors are also included. Our method significantly improves alignment, reducing the mean absolute error to 0.69m and the std to 0.84m. In all sequences, the corrected alignment consistently outperforms the GNSS-based annotations. Overall, our method improves the mean absolute alignment error by 50\% and the std by 55\% compared to the raw KITTI GNSS values.

%
%
Fig.~\ref{fig:kitti_dist00} shows the distribution of the alignment errors for the five KITTI sequences.

\noindent\textbf{Elevation Alignment on KITTI Dataset.}
We assess the effectiveness of our method in correcting vertical deviations between LiDAR point clouds and the map-based (SRTM 30m) elevation references. Fig.~\ref{fig:kitti_overall} (bottom) shows qualitative results where our method clearly improves over the original KITTI data. This improvement is evident in both the height map color representation and the cross-section comparison plots. The corrected point clouds exhibit better vertical consistency and reduced elevation artifacts. 

\begin{table}[b!]
\vspace{-7mm}
\centering
\caption{Elevation (z axis) alignment errors on the KITTI dataset.}
\label{tab:kitti_z_acc}
\vspace{-3mm}
\resizebox{0.5\textwidth}{!}{%
\begin{tabular}{c|c>{\columncolor{yellow!40}}c|c>{\columncolor{yellow!40}}c}
\hline
\multirow{2}{*}{Seq.} & \multicolumn{2}{c|}{Distance error in meters (mean$\pm$std) $\downarrow$} & \multicolumn{2}{c}{Correlation Coefficient $\uparrow$} \\
\cline{2-5}
 & KITTI Raw & After Correct. & KITTI Raw & After Correct. \\
\hline
00 & $1.42 \pm 1.70$ & $0.59 \pm 0.95$ & 0.29 & 0.69 \\
02 & $11.76 \pm 12.96$ & $4.49 \pm 5.76$ & 0.94 & 0.96 \\
05 & $0.56 \pm 0.72$ & $0.83 \pm 1.19$ & 0.77 & 0.62 \\
08 & $3.10 \pm 6.84$ & $1.16 \pm 1.92$ & 0.11 & 0.63 \\
09 & $6.77 \pm 9.35$ & $3.59 \pm 4.41$ & 0.85 & 0.94 \\
\hline
Average & $4.72 \pm 6.31$ & $2.13 \pm 2.85$ & 0.59 & 0.77 \\
\hline
\end{tabular}%
}
\vspace{-3mm}
\end{table}

\begin{figure*}
    \centering
    \includegraphics[width=1\linewidth]{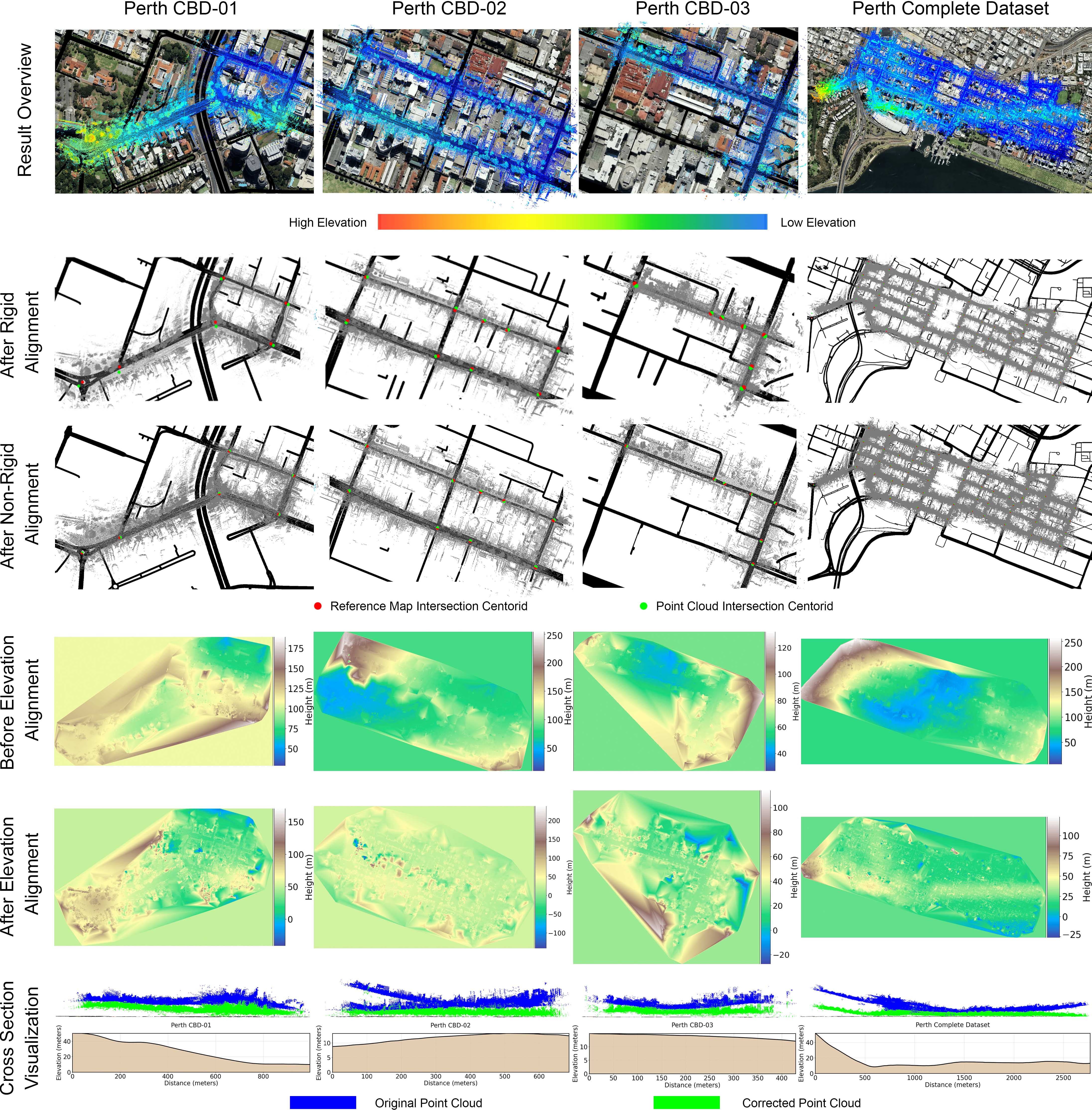}
    \caption{Qualitative results for the Perth CBD dataset. Top: point cloud alignment and map overlay for three closed loops and the complete map. Middle: comparison of intersection centroid positions before/after correction (red: map, green: point cloud). Bottom: elevation alignment results, including before/after alignment maps and cross-section profiles, showing significant reduction in global and local elevation discrepancies.}
    \label{fig:perth_overview}
    \vspace{-3mm}
\end{figure*}

Table~\ref{tab:kitti_z_acc} reports our results for elevation alignment. In Seq.~02, the mean absolute error is reduced from 11.76m to 4.49m (a 61.8\% improvement). Our method degrades the vertical alignment only for Seq.05 by 0.27m, since this sequence is mostly planar, however, the error is still well below the vertical resolution ($\pm$3.56m) of the SRTM 30m data. 
Overall, our method reduces the mean absolute vertical error for all five sequences from 4.72m to 2.13m (a 54.8\% improvement). The average correlation coefficient between the two elevation values (ours and SRTM) improves from 0.59 to 0.77 after vertical alignment correction.

%
%
Fig.~\ref{fig:kitti_z00} shows the distribution of the vertical alignment errors for the five sequences. 

\begin{table}[b!]
\vspace{-2mm}
\centering
\caption{Elevation (z axis) alignment errors on the Perth dataset.}
\label{tab:perth_z_acc}
\vspace{-3mm}
\resizebox{0.5\textwidth}{!}{%
\begin{tabular}{c|c>{\columncolor{yellow!40}}c|c>{\columncolor{yellow!40}}c}
\hline
\multirow{2}{*}{Seq.} & \multicolumn{2}{c|}{Distance error in meters (mean$\pm$std) $\downarrow$} & \multicolumn{2}{c}{Correlation Coefficient $\uparrow$} \\
\cline{2-5}
 & Before & After Correct. & Before & After Correct. \\
\hline
01 & $2.89 \pm 3.15$ & $1.45 \pm 0.82$ & 0.76 & 0.94 \\
02 & $17.84 \pm 20.15$ & $3.03 \pm 2.12$ & 0.22 & 0.81 \\
03 & $8.67 \pm 9.95$ & $1.32 \pm 1.64$ & 0.65 & 0.84 \\
Merged & $15.35 \pm 18.23$ & $3.91 \pm 4.88$ & 0.38 & 0.86 \\
\hline
\end{tabular}%
}
\vspace{-3mm}
\end{table}

\subsection{Results on the Perth CBD Dataset}
To further validate the robustness and scalability of the proposed method in GNSS-denied scenarios, we conducted experiments on the Perth CBD dataset. 


\noindent\textbf{Planimetric Alignment on Perth Dataset.}
Fig.~\ref{fig:perth_overview} presents qualitative results for three representative closed-loop regions in Perth (01, 02, 03) and the complete map. The global map-level consistency is achieved solely through map and intersection-based correction, without external positioning sensors. The middle section compares intersection centroid alignment before and after non-rigid correction. Red and green points mark map-based and point cloud based road centroids, respectively. Prior to correction, noticeable global drift and local misalignments are present. After non-rigid registration, the green and red points closely align. 

Table~\ref{tab:perth_xy_acc} summarizes improvements in planimetric alignment on Perth data. 
After non-rigid registration, the mean absolute error of the complete 3D map reduces from 5.09\,m to 2.17\,m ( 57.4\% improvement) and the std decreases from 7.27\,m to 1.07\,m (85.3\% improvement). 
For Seq.~01, we achieve the highest improvement of 88.1\% in mean absolute error (from 7.32m to 0.87m). 

\noindent\textbf{Elevation Alignment on Perth Dataset.}
The lower section of Fig.~\ref{fig:perth_overview} visualizes elevation alignment outcomes. For each region, elevation raster maps before and after correction are shown along with cross-sectional height profiles of representative road segments. Prior to correction, the LiDAR point cloud exhibits global elevation bias and local inconsistencies, seen as undulations and discontinuities when compared to terrain references. After correction, these errors are substantially reduced. The elevation profiles demonstrate strong agreement with ground truth and spatial continuity across the mapped area, confirming that vertical drift has been mitigated and geometric consistency restored—essential for 3D city modeling. 

Table~\ref{tab:perth_z_acc} reports elevation alignment metrics. Seq-02 shows the most significant improvement, where the mean absolute error reduces from 17.84\,m to 3.03\,m (83\% improvement) and the
std reduces from 20.15\,m to 2.12\,m (89.5\% improvement). For the complete 3D map, the mean absolute error reduced from 15.35\,m to 3.91\,m (74.5\% improvement) and std reduces from 18.23\,m to 4.88\,m (73.2\% improvement). We also see similar improvements in the correlation coefficient where it improves from 0.38 to 0.86 for the complete 3D map.

\vspace{-2mm}
\section{Conclusion}

We presented a post-hoc geo-registration method for terrestrial LiDAR point clouds without reliable GNSS, enabling alignment with satellite imagery. The method also corrects global biases in GNSS-referenced datasets, as demonstrated on KITTI. Through a structured pipeline, our approach aligns road structures between point clouds and satellite images, and integrates elevation data to correct vertical inconsistencies. The method is particularly suited for city-scale geo-registration tasks, where the global topological characteristics of road networks are sufficiently distinctive to resolve matching ambiguities arising from locally repetitive street patterns. The proposed evaluation metric based on the satellite road network provides a generalized and application-oriented means for assessing absolute geo-registration quality at urban scale.
Finally, we provide geo-referencing data for the publicly available complete Perth CBD dataset and the five longest sequences of the KITTI dataset to facilitate future research in this direction.


\section{Acknowledgement}
Ajmal Mian is the recipient of an Australian Research Council Future Fellowship Award (project number FT210100268) funded by the Australian Government.

\bibliographystyle{IEEEtran}  
\bibliography{ref}  

\IEEEtriggeratref{30}

\vspace{-10mm}
\begin{IEEEbiography}[{\includegraphics[width=1in,height=1.25in,clip,keepaspectratio]{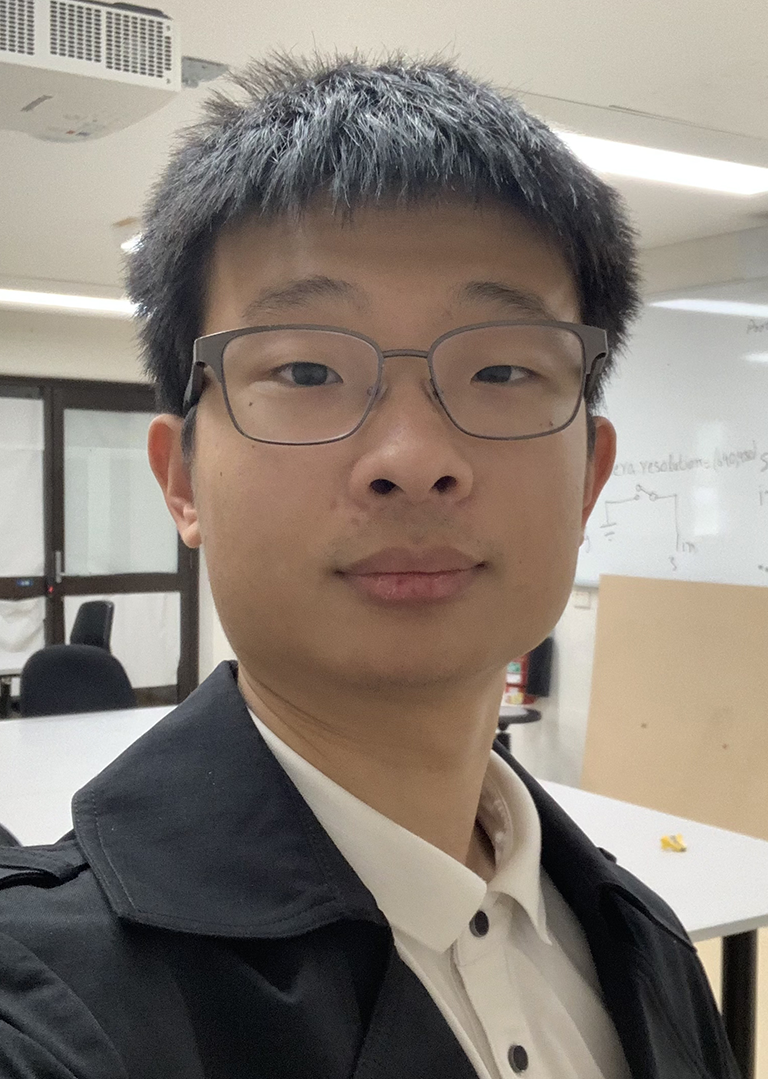}}]{Xinyu Wang}
received the B.I.T. degree in Networking and Cybersecurity from the University of South Australia, in 2018, and the B.Sc. (Hons.) degree and the Master of Information Technology degree from The University of Western Australia (UWA) in 2024. He is currently pursuing the Ph.D. degree in Computer Science at UWA, beginning in 2025. He is working as a research officer and a casual teaching staff at UWA. His research interests include 3D point cloud processing, LiDAR-based spatial modeling, geo-referencing of unstructured data, urban scene reconstruction, and geospatial machine learning.
\end{IEEEbiography}

\vspace{-10mm}

\begin{IEEEbiography}[{\includegraphics[width=1in,height=1.25in,clip,keepaspectratio]{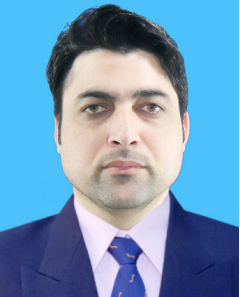}}]{Muhammad Ibrahim}
received the B.Sc. degree in Computer Systems Engineering from UET, Peshawar, Pakistan, in 2008, the M.Sc. degree in Personal Mobile and Satellite Communication from the University of Bradford, U.K., in 2010, and the Ph.D. degree in Computer Science from The University of Western Australia (UWA) in 2023. He is currently a Research Scientist at the Department of Primary Industries and Regional Development (DPIRD), WA, and an Adjunct Research Fellow with UWA. His research interests include 3D point cloud analysis, remote sensing, LiDAR-based scene understanding, precision agriculture, and deep learning for geospatial applications.
\end{IEEEbiography}

\vspace{-10mm}

\begin{IEEEbiography}[{\includegraphics[width=1in,height=1.25in,clip,keepaspectratio]{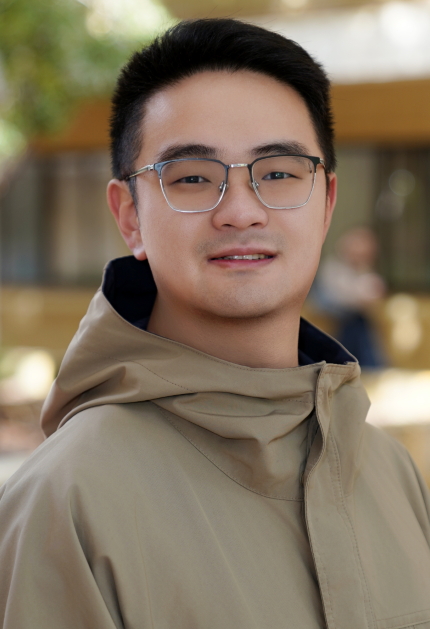}}]{Haitian Wang}
received the B.Eng. degree in Internet of Things Engineering from Northwestern Polytechnical University, Xi'an, China, in 2019, the M.Eng. degree in Computer Technology from Northwestern Polytechnical University in 2022, and the MPE (Software) degree from the University of Western Australia in 2024. He is currently a Research Scientist at the Department of Primary Industries and Regional Development (DPIRD), Western Australia, and a Research Associate at UWA. His research interests include LiDAR point cloud processing, spatial modeling, geo-registration of unstructured data, and multispectral remote sensing for precision agriculture.
\end{IEEEbiography}

\vspace{-10mm}
\begin{IEEEbiography}[{\includegraphics[width=1in,height=1.25in,clip,keepaspectratio]{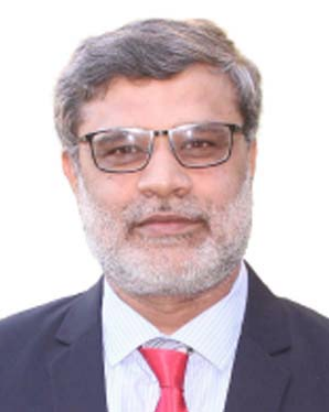}}]{Atif Mansoor}
is an Assistant Professor with the Department of Computer Science and Software Engineering, University of Western Australia. Before joining The University of Western Australia, he taught at the National University of Sciences and Technology and Punjab University in Pakistan. He was an Erasmus Mundus scholar with the Norwegian University of Science and Technology, Gjovik and Jean Monnet University, Saint-Etienne, France. He has published more than 70 papers in international conferences and journals. His research interests include computer vision, pattern classification, and cyber-physical systems.
\end{IEEEbiography}

\vspace{-10mm}

\begin{IEEEbiography}[{\includegraphics[width=1in,height=1.25in,clip,keepaspectratio]{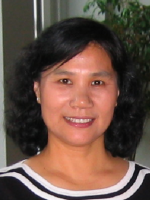}}]{Xiuping Jia} (Fellow, IEEE) received the B.Eng. degree from Beijing University of Posts and Telecommunications, Beijing, China, in 1982, and the Ph.D. degree in electrical engineering, and the Graduate Certificate in higher education from the University of New South Wales, Canberra, ACT, Australia, in 1996 and 2005, respectively. She is currently an Associate Professor at the School of Engineering Technology, The University of New South Wales. Her research interests include remote sensing, hyperspectral image processing, and spatial data analysis. She has authored or co-authored more than 280 scientific papers and a remote sensing textbook. Dr. Jia is the Editor-in-Chief of IEEE TRANSACTIONS ON GEOSCIENCE AND REMOTE SENSING.

\end{IEEEbiography}

\vspace{-10mm}

\begin{IEEEbiography}[{\includegraphics[width=1in,height=1.25in,clip,keepaspectratio]{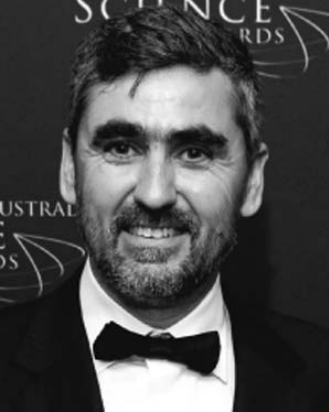}}]{Ajmal Mian} is currently a Professor of Computer Science with The University of Western Australia. He has received three esteemed national fellowships from the Australian Research Council (ARC), including the Future Fellowship award. He is a fellow of the International Association for Pattern Recognition (IAPR), an ACM Distinguished Speaker, and former President of the Australian Pattern Recognition Society. He has received several major research grants from the ARC, the National Health and Medical Research Council of Australia, Australian Department of Defense and the U.S. Department of Defense. His research interests include computer vision, machine learning, and remote sensing.  
\end{IEEEbiography}

\end{document}